\documentclass[journal]{IEEEtran}
\ifCLASSINFOpdf
\else
\fi
\usepackage{epsfig}
\usepackage{graphicx}
\usepackage{amsmath}
\usepackage{amssymb}
\usepackage{cite}

\usepackage{enumitem}
\usepackage{color, colortbl}
\usepackage[dvipsnames]{xcolor}
\usepackage{multirow}
\usepackage{changepage}
\usepackage{mathtools}
\usepackage{booktabs}

\usepackage[colorlinks]{hyperref}
\hypersetup{citecolor=blue,urlcolor=magenta}

\usepackage{subcaption} 

\definecolor{Green}{rgb}{0.82,0.82,0.82}
\definecolor{citecolor}{HTML}{0071bc}

\newcommand{\best}[1]{\textcolor{OliveGreen}{#1}}
\newcommand{\bestm}[1]{\textcolor{MidnightBlue}{\mathbf{#1}}}

\begin{document}
%
\title{Domain Adaptor Networks for Hyperspectral Image Recognition}
%
%
%

\author{Gustavo P\'erez
        and~Subhransu Maji
\thanks{G. P\'erez and S. Maji were with the Department
of Computer Science, University of Massachusetts Amherst, Amherst,
MA, 01002 USA e-mail: gperezsarabi@umass.edu.}
\thanks{}}

%
%

\markboth{} 
{Shell \MakeLowercase{\textit{et al.}}: Bare Demo of IEEEtran.cls for IEEE Journals}
%



\maketitle

\begin{abstract}
We consider the problem of adapting a network trained on three-channel color images to a hyperspectral domain with a large number of channels. 
To this end, we propose domain adaptor networks that map the input to be compatible with a network trained on large-scale color image datasets such as ImageNet.
Adaptors enable learning on small hyperspectral datasets where training a network from scratch may not be effective.
We investigate architectures and strategies for training adaptors and evaluate them on a benchmark consisting of multiple hyperspectral datasets. We find that simple schemes such as linear projection or subset selection are often the most effective, but can lead to a loss in performance in some cases. 
We also propose a novel multi-view adaptor where of the inputs are combined in an intermediate layer of the network in an order invariant manner that provides further improvements. 
We present extensive experiments by varying the number of training examples in the benchmark to characterize the accuracy and computational trade-offs offered by these adaptors.
\end{abstract}

\begin{IEEEkeywords}
convolutional neural networks, domain adaptation, transfer learning, hyperspectral images.
\end{IEEEkeywords}

%
\IEEEpeerreviewmaketitle

\section{Introduction}
\label{sec:introduction}
\begin{figure}
    \centering
    \includegraphics[width=\linewidth]{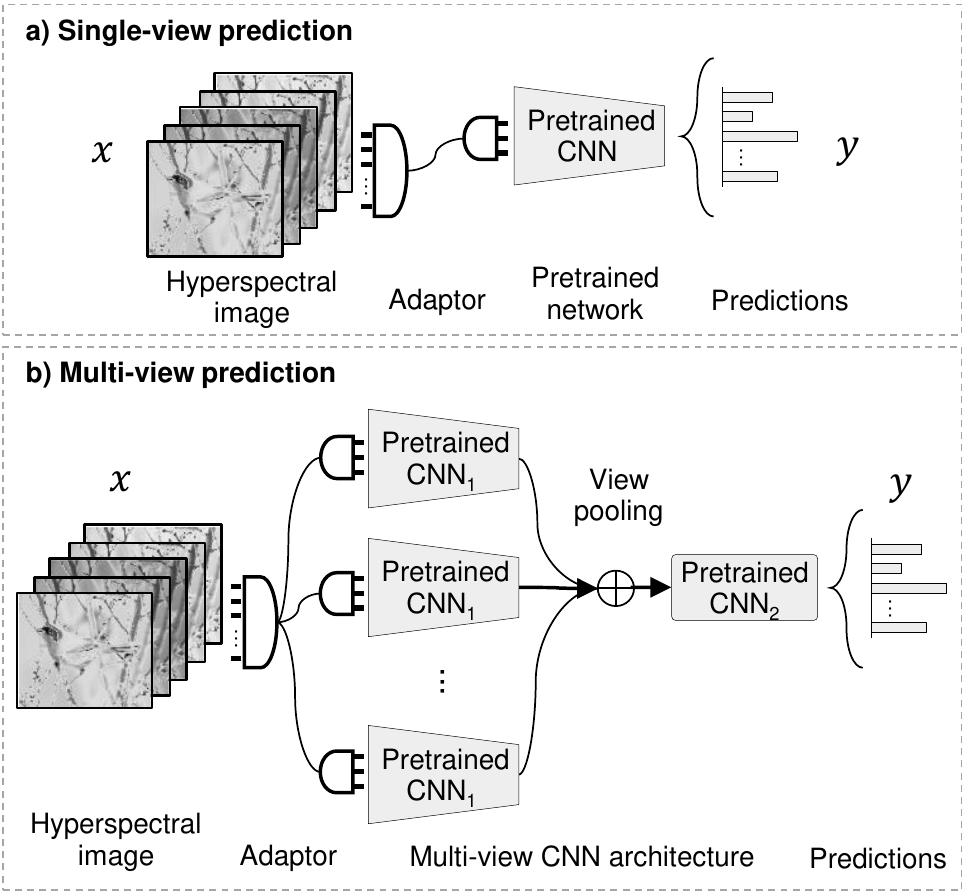}
    \caption{\textbf{Single and multi-view adaptors for hyperspectal image classification.} \textbf{(a)} An adaptor maps a multi-channel image to a three-channel image, making it compatible with a pretrained color network. \textbf{(b)} Multiple adaptors generate views of the input which are processed through a shared network to obtain the final prediction.}
    \label{fig:splash}
    \vspace{-10pt}
\end{figure}

\begin{figure*}[t]
    \centering
    \includegraphics[width=1.0\linewidth]{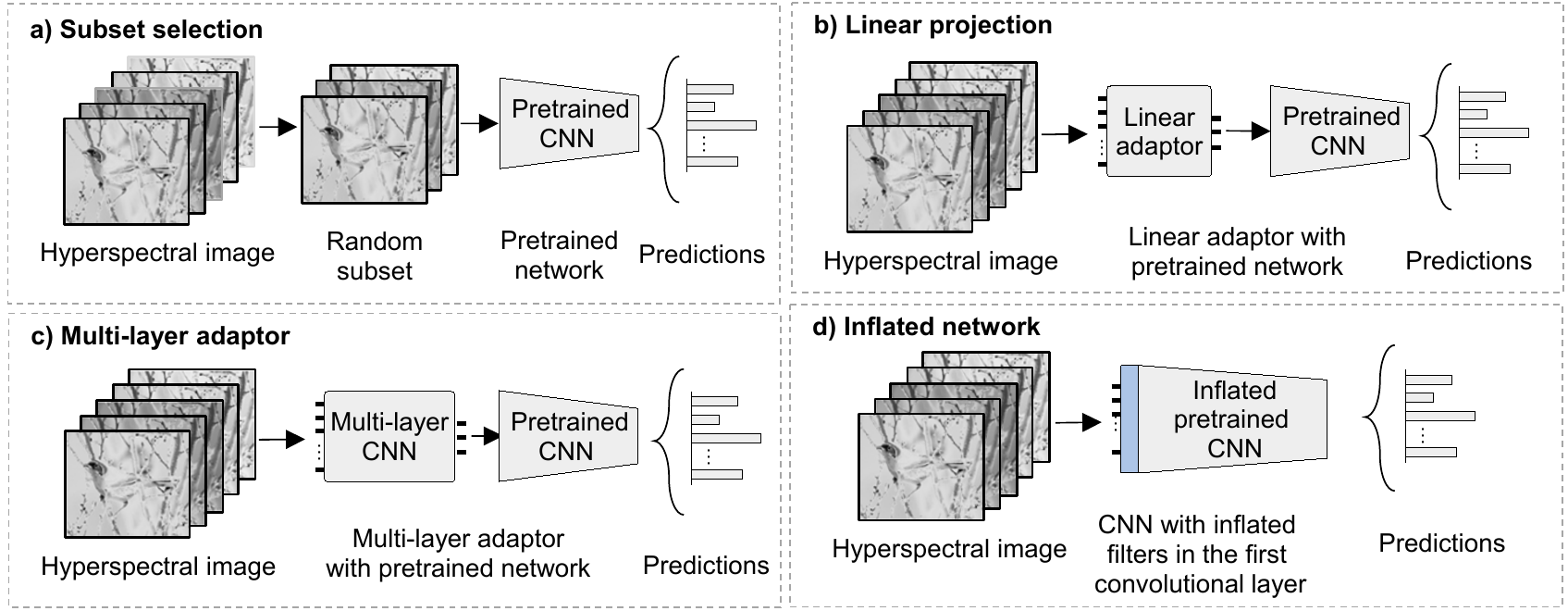}
    \caption{\textbf{Hyperspectral adaptors.} \textbf{(a)} Subset selection randomly selects a set of 3 channels. \textbf{(b)} Linear adaptor uses a linear projection from a hyperspectral image to 3 channels.  \textbf{(c)} Multi-layer adaptor uses a multi-layer network to map the input to 3 channels.  \textbf{(d)} Inflated network replaces the filters of the first convolutional layer by averaging and replicating to match the number of channels in the input.
\label{fig:methods}}
\end{figure*}

\IEEEPARstart{T}{ransferring} deep networks trained on large collections of labeled color images has been key to their success on visual recognition~\cite{transferlearning,weiss2016,zhuang2020}.
However, the effectiveness of the transfer depends in part on how related the source and target domains are.
For example, models trained on Internet images may not be as effective on recognizing medical or astronomy  images.
A further challenge arises when transferring across heterogeneous domains where some architectural modification to the network is necessary for it to process the input.
This paper studies the problem of transfer learning in this context by designing a \emph{domain adaptor network} that can be plugged in before a color image network to process hyperspectral images consisting of different number of channels than the original network was trained on.
These schemes are illustrated in Fig.~\ref{fig:splash} and Fig.~\ref{fig:methods}.

Our problem is motivated by the fact that hyperspectral domains lack pretrained networks that can serve as general-purpose feature extractors.
Thus one might benefit from architectural innovations and pre-training in the domain of color images which are readily available in ``model zoos" in modern libraries.
While the literature contains several schemes for transferring pretrained networks from color to hyperspectral domains, a systematic evaluation is lacking.
Our \emph{first contribution} is therefore a benchmark of six hyperspectral datasets (\S~\ref{sec:datasets}) divided into two groups as seen in Fig.~\ref{fig:datasets}.
The first group contains three remote sensing datasets: LEGUS~\cite{legus}, So2Sat LCZ42~\cite{so2sa}, and EuroSAT~\cite{eurosat}. 
The second group modifies the images in Caltech-UCSD birds~\cite{cub}, FGVC aircraft \cite{aircrafts}, and Stanford cars~\cite{cars} datasets by synthetically expanding the channels.
The synthetic datasets allow us to measure the effectiveness of transfer by comparing it with the performance on the unmodified color images and control the amount of domain shift by varying the number of channels.
Despite their simplicity these datasets reveal some difficulties in transfer learning. For example, permuting the color channels makes the transfer significantly less effective on the CUB dataset. We analyze this phenomenon and it's implications in \S~\ref{sec:domain_shift}.

Our \emph{second contribution} is an evaluation of common schemes in the literature. As seen in Fig.~\ref{fig:methods} these include: (a) a linear projection; (b) selecting a subset; (c) a multi-layer network; (d) ``inflating" the first layer of the network. 
We also investigate techniques for pre-training the adaptors in an unsupervised manner.
Finally, we propose a novel multi-view scheme that generates a prediction by aggregating information across ``views" of the input as illustrated in Fig.~\ref{fig:splash}.

Our experiments suggest that simple schemes such as linear projection or subset selection are often the most effective (Tab.~\ref{table:expvgg} \& Fig.~\ref{fig:violin}). 
Deeper adaptors and unsupervised training of adaptors offer little additional benefit highlighting the difficulty of adapting a network layer far from the output layer during transfer learning.
Multi-view adaptors can be used with any baseline adaptor and provide consistent improvements when the domain has a large number of channels (Tab.~\ref{table:expvgg}).
The multi-view adaptor can be thought of a light-weight ensemble, as these networks require few additional parameters compared to a separate network trained on each view.
More importantly, all of these schemes are significantly better than training a custom network from scratch underlying the need to study techniques to adapt color image networks to heterogeneous domains.

In summary our contributions are:

\begin{enumerate}
\setlength\itemsep{0.1em}
    \item   We propose a benchmark for investigating the effectiveness of  transfer learning on hyperspectral images (\S~\ref{sec:datasets}). 
    We analyze the effectiveness of adaptor networks that map multi-channels to a three-channel image compatible pretrained network (\S~\ref{sec:adaptors} and \S~\ref{sec:experiments}).
    \item We propose a novel multi-view scheme that is significantly more lightweight than ensembles but offer consistent gains. On the technical side, we propose a regularization scheme that encourages diversity across views which provides further benefits (\S~\ref{sec:multiview}).
    \item We illustrate the difficulty in transfer learning even when domain shifts are simple (e.g., color channels are permuted), suggesting the need to investigate techniques beyond fine-tuning, such as self-supervised and semi-supervised learning which can better exploit domain-specific data (\S~\ref{sec:experiments}). 
\end{enumerate}

The training and evaluation source code is publicly available at
\href{https://github.com/gperezs/hyperspectral\_domain\_adaptors}{https://github.com/gperezs/hyperspectral\_domain\_adaptors}.
\section{Related work}
\label{sec:related_work}
\subsection{Transfer learning}
Transfer learning is an effective strategy for learning from a few examples.
Within the context of deep networks, this is typically done by training a network on a large labeled dataset (e.g., ImageNet~\cite{imagenet}) and fine-tuning its parameters on the downstream task after adding task-specific layers.
Schemes vary from training a subset of layers~\cite{transferlearning}, or parameters (e.g., batch norm statistics~\cite{bn}), to adding a secondary network to parameterize the changes~\cite{houlsby2019} and custom architectures where parts of the network are initialized (e.g., B-CNN~\cite{bcnn}, Faster R-CNN~\cite{fasterrcnn}, etc.).
For robustness to domain shifts, such as changes in image resolution, a number of techniques based on the aligning the distribution statistics of deep network features across domains have been proposed~\cite{long2017,herath2019,wang2020}.
When paired data across domains is available one might learn the mapping directly (e.g., using a GAN~\cite{goodfellow2014generative}) or apply cross-modal training techniques (e.g., CQD~\cite{cross-quality-distillation}, data distillation~\cite{radosavovic2017}) to improve transfer. 
However, this requirement is rarely met in practice.
While most techniques have been studied when the source and target domains have the same structure, there are cases when transfer learning across heterogeneous domains is effective.
For example, recognizing 3D shapes by rendering them as images, or classifying audio by rendering them as spectrograms. Our work aims to design such adaptors for hyperspectral domains.

\subsection{Hyperspectral image datasets}
Hyperspectral images are common in remote sensing (e.g., RADARs, Satellites, Telescopes) and medical imaging (e.g.,\cite{medimg,Ferris2001}) as they reveal properties not easily seen in color images. 
For instance, remote sensing data acquired from earth observation satellites are routinely used for land cover classification, infrastructure planning, and population assessment~\cite{LC1,LC2,LC3,LC4,zhu2017,qiu2020}. 
The Hubble Space Telescope (HST) captures data through a variety of filters, each passing specific wavelengths of light, which are used in various scientific analysis.
Tasks in these domains often lack large labeled datasets to train deep networks and transfer learning from color images provides a compelling alternative. However, the large differences in domains and their heterogeneous structure poses a barrier.

\subsection{Hyperspectral image classification} 
There is a large literature on applying classical learning methods such as decision trees, random forests, and support vector machines for hyperspectral data such as on earth monitoring with satellite imagery~\cite{camps2018,Ferris2001}. 
Deep networks can be also be used in a straightforward manner when training a network from \emph{scratch} and have been successfully applied to different tasks~\cite{barchi2020, morpheus, khalifa2017, zhu2019, martin2019, mittal2019}. 
However, the lack of large-scale labeled datasets makes training from scratch less effective than say transfer learning via color image networks.
Several schemes are possible to account for the different number of channels. 
A common strategy is to combine color networks each trained on a different subset of channels. 
While this has been applied successfully for some tasks (e.g., combining flow and appearance for video~\cite{simonyan2014}, or classifying astronomy images~\cite{wei2020}), strategies for selecting subsets and combining them vary.
Ensembles also increase the computational complexity and model size.
Another approach is to \emph{manually select} a subset of channels based on domain knowledge \cite{Bialopetravicius2019,Bialopetravicius2020}. 
The \emph{dimensionality reduction} can also be learned using PCA~\cite{pca}. While this is often used as a visualization tool, it is also been shown to be effective for transfer (e.g., classifying RADAR data~\cite{mishra2006}).
Another approach is based on \emph{inflating} the network by replicating the filters in the first layer. This strategy has been effectively applied to transfer color network to spatio-temporal~\cite{inflated,huang2020} and 3D domains~\cite{lalonde2019}.
Even when the number of channels are fewer, one can benefit from expanding it to a color image. 
For example, depth data rendered as color images (e.g., HHA encoding~\cite{gupta2014}) has been shown to lead to a better performance with pretrained color networks.


\section{A hyperspectral classification benchmark}
\label{sec:datasets}
\begin{figure*}[t]
    \centering
    \includegraphics[width=1.0\linewidth]{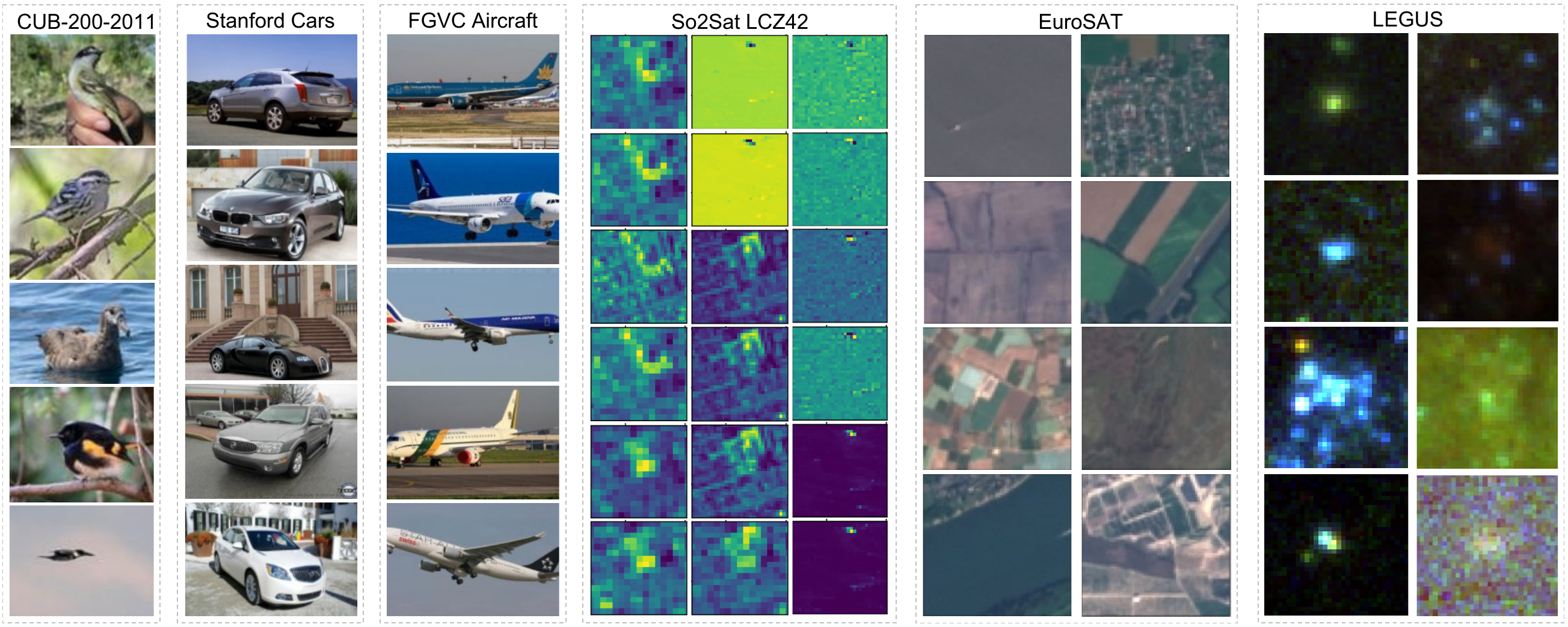}
    \caption{\textbf{Datasets.} Sample images of the six datasets in the proposed benchmark. 
    In the case of synthetic datasets (CUB-200-2011, Stanford Cars, and FGVC Aircraft) we show the original images. For So2Sat LCZ42, we show the 18 bands of a single sample. For EuroSAT, we show the RGB version of the images also publicly available. For LEGUS, we show RGB images using the conversion described at the end of \S~\ref{sec:domain_shift}.  
    \label{fig:datasets}}
\end{figure*}

Our benchmark consists of six classification datasets with varying number of channels and classes as illustrated in Tab.~\ref{table:datasets} and Fig.~\ref{fig:datasets}.
The datasets are divided in two groups called \emph{synthetic} and \emph{realistic}. 
The synthetic group consists of RGB images synthetically expanded to 5 or 15 channels, while the realistic group consists of hyperspectral datasets from remote sensing applications.
They are described in detail next.

\begin{table}
\centering
\begin{tabular}{l c c c c}
\toprule
Dataset  &  Train & Test & Classes & Channels\\
\hline
CUB-200-2011 &	            5994 &	    5794 &		200     & \{5, 15\}\\
Stanford Cars &		        8144 &		8041 &		196     & \{5, 15\}\\
FGVC Aircraft &		        6667 &		3333 &		100     & \{5, 15\}\\
\hline
So2Sat LCZ42 &		        10000 &     10000 &		17      & 18     \\
EuroSAT  &		            10000 &		17000 &		10      & 13    \\
LEGUS  &		            12376 &		3095 &		4       & 5     \\
\bottomrule\\
\end{tabular}
\caption{\textbf{Datasets used in our benchmark}. The table shows the number of training samples, testing samples, classes, and the number of channels in the each dataset. The first three datasets are synthetically expanded to 5 or 15 channels. 
\label{table:datasets}}
\end{table}

\subsection{Synthetic datasets} 
We generate synthetic versions of the RGB images by mapping each color $\in \mathbb{R}^3$ to a vector in $\mathbb{R}^k$ with $k > 3$. We use $k$-means to cluster the color values of images in each dataset to find $k$ centers. Then each channel is computed as:  
\begin{equation}
    f[x,y,i] = \exp\big(-||f_0[x,y] - c_i||_2^2\big), 
\end{equation}
where $f_0[x,y] \in \mathbb{R}^3$ is the RGB value of the pixel at location $(x,y)$, and $c_i \in \mathbb{R}^3$ is the $i^{th}$ center calculated with $k$-means. 
This produces a $k$-channel image. We now describe the synthetic datasets.

\begin{itemize}[leftmargin=0.15in]
\setlength\itemsep{0.1em}
    \item \textbf{CUB-5 and CUB-15.} The CUB-200-2011~\cite{cub} dataset contains 11788 images divided in 5994 images for training and 5794 images for testing. The dataset contains images of 200 bird species. We generate the CUB-5 and CUB-15 dataset by expanding each color image to 5 and 15 channels respectively.

    \item \textbf{Cars-5 and Cars-15.}
    The Stanford cars~\cite{cars} dataset contains 16185 images of 196 car models. The data is split into 8144 training images and 8041 testing images, where each class has been divided following a 50-50 split. 
    As before, we generate the Cars-5 and Cars-15 dataset by expanding the channels.

    \item \textbf{Aircraft-5 and Aircraft-15.}
    The FGVC aircraft~\cite{aircrafts} dataset contains 10000 images of aircraft, with 100 images for each of 100 different aircraft model variants. The benchmark proposes a train-validation-test split of roughly the same number of images. In all our experiments, we use only the image-level variant annotations and the trainval-test splits of 6667 and 3333 images respectively. Aircraft-5 and Aircraft-15 were generated in a similar manner to the other synthetic datasets.
\end{itemize}

\subsection{Realistic datasets}
We select datasets with images with more than 3 color channels. 
We choose datasets with different domain shifts from natural images. In addition, we include a reduced version of these datasets with only 1000 training images (So2Sat$^S$, EuroSAT$^S$, and LEGUS$^S$) to study the adaptor networks when trained with less samples.
The realistic datasets are described below.

\begin{itemize}[leftmargin=0.15in]
\setlength\itemsep{0.1em}
    \item \textbf{So2Sat LCZ42.} 
    The Social Media to EO Satellites Local Climate Zones 42 (So2Sat LCZ42)~\cite{so2sa} is a dataset consisting of co-registered synthetic aperture radar and multispectral optical image patches acquired by the Sentinel-1 and Sentinel-2 remote sensing satellites. The task is to classify the images into 1 of 17 local climate zones (LCZ). For the complete version of the dataset we use a train-test split of 10000 images with 18 spectral bands each. 

    \item \textbf{EuroSAT.}
    The Dataset for Land Use and Land Cover Classification (EuroSAT)~\cite{eurosat} is based on Sentinel-2 satellite images covering 13 spectral bands. The dataset consists of 27000 labeled and geo-referenced images with annotations of 10 classes of terrain. 

    \item \textbf{LEGUS.}
    The Legacy ExtraGalactic UV Survey (LEGUS)~\cite{legus} consists of 50 galaxies at distances between 3.5~Mpc and 16~Mpc, 
    HST in five bands spanning the ultraviolet to infrared spectrum.
    The task is to classify the star-clusters within each galaxy into four classes.
    The first three classes correspond to different morphological properties and class 4 refers to non-cluster objects like foreground stars, background galaxies, or artifacts. 
\end{itemize}

\section{Adaptors for hyperspectral images}
\label{sec:adaptors}
The architecture of a pretrained color network has to be modified to process a multi-channel image. Fig.~\ref{fig:methods} illustrates the four choices of adaptor architectures.

\begin{itemize}[leftmargin=0.15in]
\setlength\itemsep{0.1em}
    \item \textbf{Linear projection.}
    A linear adaptor maps a hyperspectral image $x \in \mathbb{R}^{n \times m \times k}$ to a view $ v(x) \in \mathbb{R}^{n \times m \times 3}$ with a linear projection from $k$ channels. The projection is implemented as a single convolutional layer with 3 filters of size $1 \times 1\times k$. 
    These are initialized randomly or using PCA, and jointly trained with the network during fine-tuning.

    \item \textbf{Subset selection.} The mapping is achieved by selecting three channels out of the $k$ input channels to obtain a view $v(x) \in \mathbb{R}^{n \times m \times 3}$. 
    We select the channels randomly.
  
    \item \textbf{Multi-layer adaptor.} A multi-layer adaptor is a multi-layer neural network that maps the hyperspectral image $ x \in \mathbb{R}^{n \times m \times k}$ to a view $v(x) \in \mathbb{R}^{n \times m \times 3}$. The adaptor is attached to the pretrained network and trained jointly from scratch or initialized using unsupervised training.

    \item \textbf{Inflated network.} We inflate the network by replacing the filter weights of the first convolutional layer by first averaging the filter weights across the channel dimension and replicating it as many times as the number of channels in the input. 
    Let $\theta \in \mathbb{R}^{h \times w \times 3 \times n}$ be the $n$ filters 
    in the first layer of network.
    This is inflated to parameters $\hat{\theta} \in \mathbb{R}^{h \times w \times k \times n}$ with each filter $\hat{\theta}^j$ given by:
    \[ \hat{\theta}^j=
      \underbrace{%
        \begin{pmatrix}
            \text{mean}(\theta^j) &  \text{mean}(\theta^j) & \cdots & \text{mean}(\theta^j)
        \end{pmatrix}%
       }_{k \text{ times}}
    \]
\end{itemize}

\subsection{Multi-view adaptors}
\label{sec:multiview}
All adaptors except the inflated network reduce the number of channels before feeding it to the pretrained network, which may lead to a loss of information.
An alternate strategy is to train an ensemble, where different models are trained on a subset of channels and then combined to produce the final prediction. 
However this scheme leads to a increase in the parameters and also impacts the speed of classification.

To address this issue we propose a \emph{novel} multi-view scheme that combines information from multiple views of the hyperspectral image using a shared network.
Each view is obtained using a different adaptor (e.g., different subsets, or different linear projections). 
Instead of training a separate network, the multi-view scheme passes each view through the first section of the network (CNN$_1$ in Fig. \ref{fig:splash}). Then the activations are aggregated using a \emph{view pooling layer} using an orderless set aggregation scheme.
Finally, the aggregated activations are passed through the remaining section of the network (CNN$_2$ in Fig. \ref{fig:splash}) to produce the final output. 
The scheme allows the view pooling layer to be added at any layer in the network.
Formally a set of views $v_i(x) \in \mathbb{R}^{n \times m \times 3}$, $i=1,\ldots,N$ are obtained from the input $x$ using any adaptor scheme described earlier. 
The output of the multi-view model is 
\begin{equation}
y = \text{CNN}_{2}\Big(\Phi\big\{\text{CNN}_{1}(v_i(x))\big\}\Big), i=1,\ldots,N, 
\end{equation}
where $\Phi\{\cdot\}$ is an orderless set aggregation function such as the mean or max. We use max in our experiments.

Unlike ensembles we share parameters across views --- the only view specific parameters are in the adaptor layer which are significantly fewer than the number of network parameters ($< 0.001\%$). 
The memory and computational cost are increased depending on which layer is view pooled. For example early layer view-pooling leads to smallest increase in cost. While last layer view pooling is as slow as an ensemble, but still leads to parameter savings. We present these details in \S~\ref{sec:implementationdetails}.
Multi-view networks exploit the ability of deep networks to combine multiple sources of information which has been exploited for other tasks such as 3D shape classification from multiple views~\cite{multi-view}.

\begin{figure}[t]
\centering
\includegraphics[width=1.0\linewidth]{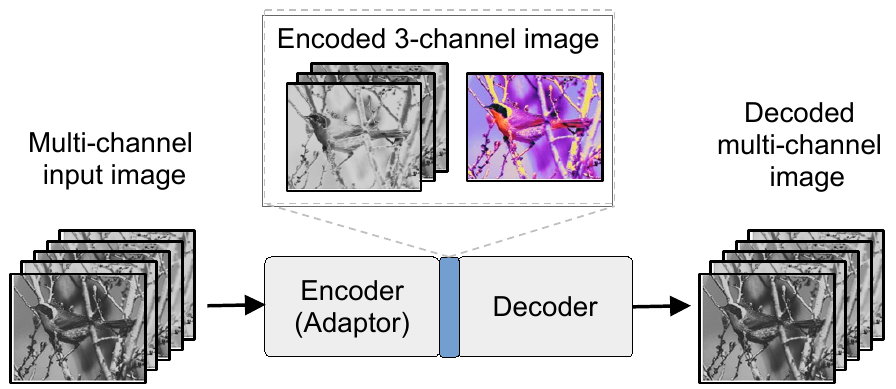}
\caption{\textbf{Multi-layer adaptor pretraining} as an auto-encoder with a 3-channel intermediate representation. 
\label{fig:autoencoder}}
\end{figure}

\section{Experiments}
\label{sec:experiments}
We first describe our experimental setup and implementation details (\S~\ref{sec:implementationdetails}). 
Then we present our results across datasets in the benchmark comparing the performance of different single-view and multi-view adaptors, and ablation studies varying the number of views and initialization schemes (\S~\ref{sec:results}).
Finally we analyze the limits of adaptors by analyzing the effect of domain shifts on the synthetic datasets (\S~\ref{sec:domain_shift}).


\begin{table*}[t]
\renewcommand{\arraystretch}{1.2}
\centering
\begin{tabular}{l c|c c c c|c c c c}
  \multicolumn{10}{c}{\textbf{Performance using VGG-D network}}\\
\toprule
& & \multicolumn{4}{c|}{Single-view} &  \multicolumn{4}{c}{Multi-view} \\
        \cline{3-10}
          & From & Linear & Inflated & Multi-layer &  Subset   & \multicolumn{2}{c}{Subset selection} & \multicolumn{2}{c}{Linear adaptor}      \\
Dataset    & scratch & adaptor & network & adaptor & selection & 2 & 5 & 2 & 5 \\
\hline
\emph{Synthetic}\\
\hline
CUB-5       & $0.5\pm0.0$ & $38.4\pm3.1$ & $30.5\pm1.1$ & $41.3\pm2.8$ & $\best{47.1\pm4.0}$  
            & $45.0\pm1.7$ & $49.8\pm0.5$ & $52.7\pm0.6$ & $\bestm{55.1\pm0.3}$  \\
CUB-15      & $0.5\pm0.0$ & $41.2\pm0.9$ & $26.0\pm2.8$ & $\best{45.4\pm5.1}$ & $\best{45.9\pm5.0}$  
            & $49.9\pm2.7$ & $\bestm{56.5\pm1.0}$ & $\bestm{57.3\pm0.5}$ & $\bestm{56.7\pm0.9}$  \\
\hline
Cars-5      & $2.2\pm0.8$ & $70.5\pm2.1$ & $70.2\pm1.4$ & $\best{74.4\pm1.7}$ & $\best{72.6\pm2.0}$  
            & $75.4\pm0.9$ & $\bestm{76.7\pm0.5}$ & $\bestm{76.7\pm0.4}$ & $\bestm{76.8\pm0.5}$   \\
Cars-15     & $2.7\pm0.4$ & $\best{70.7\pm3.3}$ & $68.7\pm3.4$ & $\best{70.1\pm2.3}$ & $\best{72.2\pm1.7}$  
            & $75.8\pm0.4$ & $77.1\pm0.3$ & $77.1\pm0.2$ & $\bestm{78.1\pm0.4}$   \\
 \hline
Aircraft-5  &  $1.0\pm0.0$ & $\best{79.6\pm0.8}$ & $76.6\pm0.9$ & $\best{79.5\pm0.5}$ & $\best{79.2\pm1.0}$  
            & $\bestm{80.8\pm0.9}$ & $\bestm{81.5\pm0.3}$ & $81.1\pm0.4$ & $\bestm{81.8\pm0.2}$   \\
Aircraft-15 &  $1.0\pm0.0$ & $78.4\pm0.6$ & $77.6\pm0.9$ & $79.1\pm1.2$ & $\bestm{79.8\pm2.2}$  
            & $\bestm{81.3\pm1.0}$ & $\bestm{81.6\pm0.5}$ & $\bestm{81.4\pm0.5}$ & $\bestm{81.3\pm0.4}$    \\
\hline
\emph{Realistic}\\
\hline
So2Sat     & $50.3\pm0.8$ & $\best{52.8\pm0.3}$ & $51.8\pm0.3$ & $\best{52.8\pm0.7}$ & $\best{51.6\pm3.5}$ 
          & $55.6\pm1.0$ & $\bestm{57.4\pm0.4}$ & $\bestm{57.0\pm0.4}$ & $\bestm{57.2\pm0.5}$    \\
So2Sat$^S$ & $34.5\pm0.7$ & $\best{45.1\pm0.5}$ & $40.8\pm0.8$ & $40.8\pm3.1$ & $\best{43.8\pm4.9}$  
          & $49.8\pm1.2$ & $\bestm{51.4\pm0.3}$ & $45.2\pm0.7$ & $48.6\pm0.8$   \\
\hline
EuroSAT     & $95.7\pm0.2$ & $97.1\pm0.3$ & $96.7\pm0.3$ & $\best{97.7\pm0.2}$ & $96.6\pm0.7$  
            & $\bestm{97.4\pm0.4}$ & $\bestm{97.7\pm0.2}$ & $\bestm{97.4\pm0.2}$ & $\bestm{97.5\pm0.2}$   \\
EuroSAT$^S$ & $75.2\pm2.6$ & $88.4\pm0.5$ & $83.6\pm1.8$ & $\best{92.8\pm0.7}$ & $86.6\pm3.6$  
            & $91.6\pm2.2$ & $93.7\pm0.5$ & $\bestm{94.4\pm0.2}$ & $\bestm{94.4\pm0.2}$  \\
\hline
LEGUS       & $51.9\pm0.5$ & $\best{63.2\pm0.5}$ & $61.4\pm0.4$ & $\best{62.8\pm0.3}$ & $61.1\pm1.4$  
            & $64.2\pm0.4$ & $\bestm{65.3\pm0.5}$ & $64.2\pm0.4$ & $63.9\pm0.6$   \\
LEGUS$^S$   & $27.2\pm5.5$ & $\best{54.9\pm0.8}$ & $51.8\pm1.3$ & $\best{53.1\pm1.9}$ & $\best{51.3\pm2.9}$  
            & $55.0\pm0.6$ & $\bestm{59.3\pm0.6}$ & $55.7\pm1.8$ & $\bestm{60.0\pm1.6}$    \\
 \bottomrule
  \multicolumn{5}{l}{\footnotesize $^S$ Smaller version of the dataset using 1000 training samples.}\\
\end{tabular}
\caption{\textbf{Results using hyperspectral domain adaptors.} Accuracy (\%) for single and multi-view adaptors on benchmark consists of the six datasets using a ImageNet pretrained VGG-D network. 
The top group indicates the \emph{synthetic} datasets while the bottom group represents \emph{realistic} datasets. 
The best results using single-view adaptor networks are shown in green color. The best overall results are shown in bold blue color.
Fig.~\ref{fig:violin} shows the same results in a different format.
Results using ResNet18 and ResNet50 are in supplementary material where the trends are similar.
\label{table:expvgg}}
\end{table*}

\begin{figure*}
    \centering
    \begin{subfigure}[t]{0.448\textwidth}
        \raisebox{-\height}{\includegraphics[width=\textwidth]{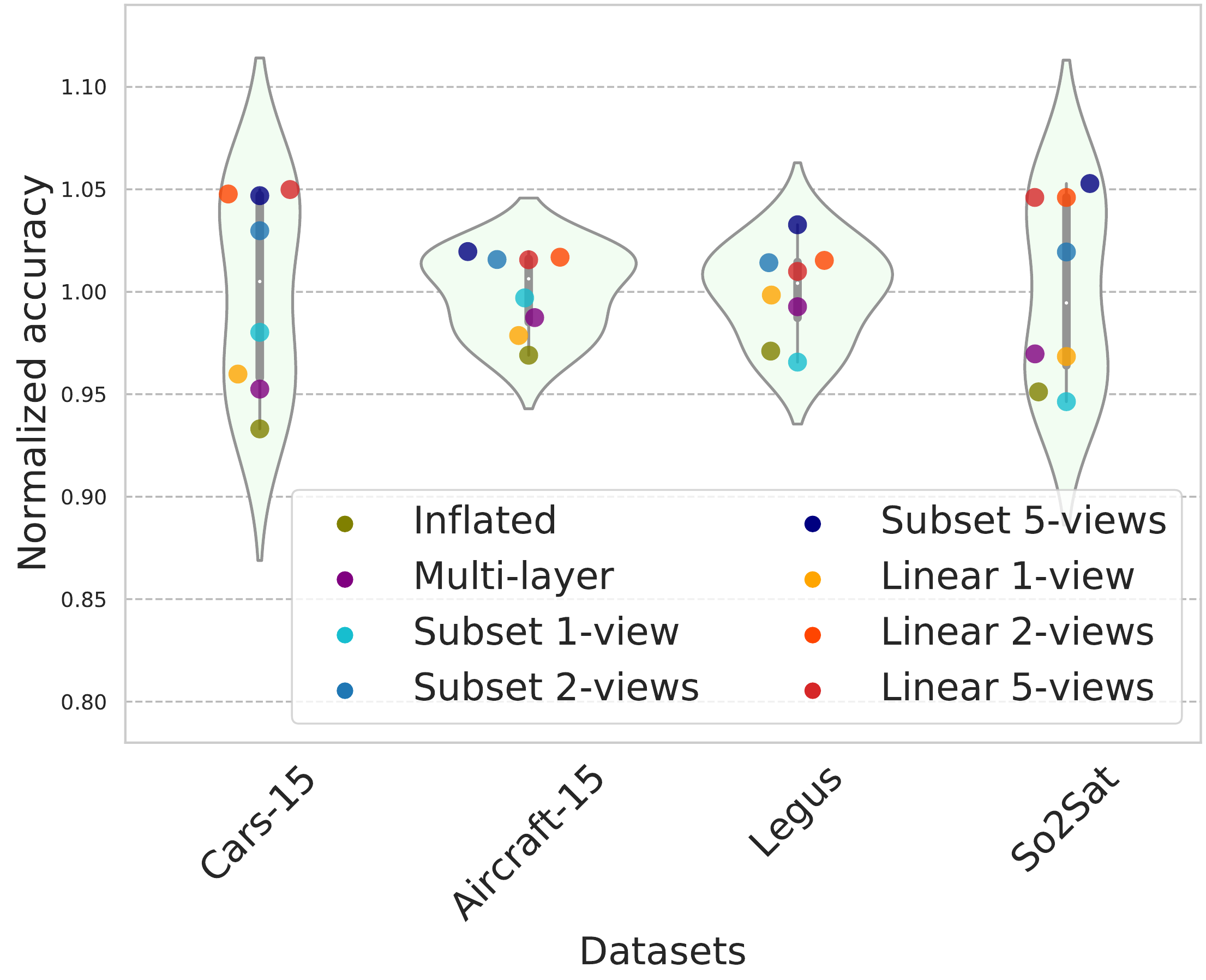}}
    \end{subfigure}
    \hfill
    \begin{subfigure}[t]{0.532\textwidth}
        \raisebox{-\height}{\includegraphics[width=\textwidth]{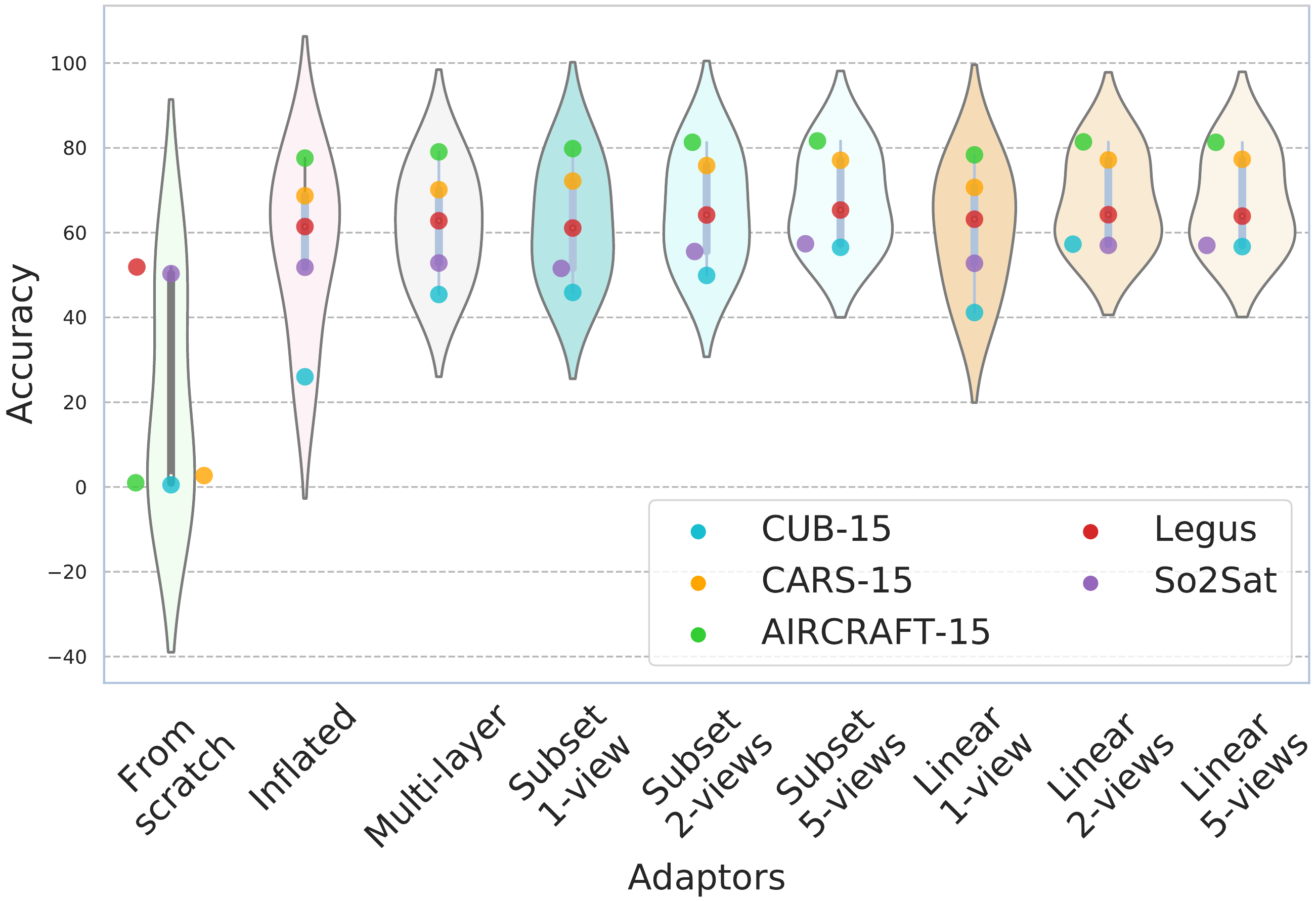}}%
    \end{subfigure}
    \caption{\textbf{Performance on the benchmark.} Distribution of accuracy for each dataset (\emph{left}) and each method (\emph{right}) using a pre-trained VGG-D network. The accuracy on the left plot is divided by the mean accuracy for better visualization. Adaptors are effective at transfer compared to training from scratch and multiple views improve performance (groups of blue and orange violin plots on the right). \emph{Best viewed in color.} }
    \label{fig:violin}
\end{figure*}

\subsection{Implementation details}
\label{sec:implementationdetails}
\noindent
\textbf{Fine-tuning details.} We use an ImageNet pretrained VGG-D~\cite{vgg}, ResNet18, and ResNet50~\cite{resnet} networks in our experiments. 
For transfer learning we replace the last fully-connected layer to match the number of classes of each dataset and train the network jointly with the input adaptor using Adam~\cite{adam} optimizer to minimize a multi-class cross-entropy loss on the image-level annotations using the train-test splits as described in \S~\ref{sec:datasets}. 
We run our experiments with synthetic datasets for 30 epochs using a learning rate of 1E-04, a batch size of 64 images, and data augmentation with horizontal flips.
We decrease the learning rate by a factor of 10 every 7, 15, and 10 epochs for birds, cars, and aircraft datasets respectively.
For realistic datasets, we train the models for 10, 6, and 15 epochs for So2Sat, EuroSAT, and LEGUS respectively using a learning rate of 1E-04 (with a decrease by a factor of 10 every 5 epochs for LEGUS, and 4 epochs for EuroSAT and So2Sat). We use a batch size of 32 images and data augmentation with random flips. In particular for LEGUS, which exhibit rotational invariance, we use random rotations.

\noindent
\textbf{Training from scratch.} For each dataset we train the three networks from scratch by modifying the first layer to match the number of input channels and the last layer to match the number of classes of each dataset. We run the experiments using 17 times more epochs compared to fine-tuning based on~\cite{kornblith2019}. We use a learning rate of 1E-04, a batch size of 64 images, and data augmentation with horizontal flips. We decrease the learning rate by a factor of 10 every 14, 30, and 20 epochs for birds, cars, and aircraft datasets respectively, and every 10 epochs for LEGUS and 8 epochs for So2Sat and EuroSAT. 

\noindent
\textbf{Multi-layer adaptor architecture.}
The multi-layer adaptor is a network with 4 convolutional layers, each one with 16 filters of size $3\times 3$, batch normalization, and ReLU activations with no stride or pooling layers. The parameters are either initialized randomly or learned as an auto-encoder using the training samples of each dataset. When pretraining the adaptor, we use Adam optimizer to minimize the mean squared error between the prediction and the hyperspectral input image. We train the adaptor for 10 epochs using a learning rate of 1E-3 and a batch size of 64 images.

\noindent
\textbf{Multi-view adaptors.}
We perform experiments using 2 and 5 views per image. 
Also, in our multi-view setup, the effective batch size is smaller since we are using shared parameters across all the views, so we apply a linear scaling rule (as proposed in \cite{goyal}) to the learning rate, learning rate schedule, and the number of training epochs. 
We place the view pooling layer at conv$_{5\_3}$ layer in the VGG-D, conv$_{5\_2}$ in the ResNet18, and conv$_{5\_3}$ in the ResNet50 network.
We report results of the multi-view experiments using linear projection and subset selection.
For datasets with large number of channels we also experiment with larger number of views.

\noindent
\textbf{Adaptor training.}
During training we found it helpful to increase the learning rate of the linear and multi-layer adaptors by a factor of 10 compared to the rest of the network. Without this the weights do not change from their initial value. 
We obtain an average of 3.7\% increase in performance by increasing the learning of the adaptors for the synthetic datasets, and an average of 5.8\% by increasing the learning of the adaptors for the realistic datasets.

\noindent
\textbf{Diversity regularization.} 
In the case of multi-view with learnable adaptors (e.g. linear projection), we propose a regularization to enforce the adaptors to learn to produce independent views from each other with the goal of increasing the amount of information per number of views. Formally, given the $k$ views $v(x) \in \mathbb{R}^{n \times m \times 3k}$ produced by a multi-view adaptor, its gram matrix $G \in \mathbb{R}^{3k\times 3k}$ is produced by $G=\hat{v}\hat{v}^T$, where $\hat{v}\in \mathbb{R}^{3k\times nm}$ is a reshaped version of the views. Our regularization for a batch of $N$ images is given by: 
\begin{equation}
    R=\alpha\sum_i^N||G_i||_p
\end{equation}
where $p$ is the order of the norm and $\alpha$ is a scaling factor. In our experiments we use the spectral norm ($p=2$) and a scaling factor of $\alpha=\text{1E-2}$. The regularization leads to an average increase of 1.3\% for synthetic datasets and 1.1\% for realistic datasets when used with multi-view linear adaptors. All our multi-view linear adaptor results are obtained using this regularization. 

\noindent
\textbf{Baseline on synthetic datasets.}
For the synthetic datasets we fine-tune a pretrained network using the original RGB images. This provides a reference upper bound on the performance the model can achieve and is shown in Tab.~\ref{table:sfinetune}. 

\begin{table}
\centering
\begin{tabular}{l c c c c }
\toprule
& &  \multicolumn{3}{c}{Accuracy (\%)} \\
Dataset  & & VGG-D & ResNet18 & ResNet50 \\
\hline
CUB &	     &       72.0 &	70.5 & 77.0 \\ 
Stanford Cars &		  &      78.2 & 80.6 & 86.7 \\ 
FGVC Aircraft &		   &     81.1 &	76.9 & 84.0 \\
\bottomrule
\end{tabular}
\caption{\textbf{Performance on the synthetic datasets.} Accuracy by fine-tuning an ImageNet pretrained network on the RGB images of each dataset. These provide an upper-bound on the performance of the synthetic datasets. 
\label{table:sfinetune}}
\end{table}

\subsection{Results and discussion}
\label{sec:results}
Tab.~\ref{table:expvgg} shows the results on all datasets using VGG-D. 
In Fig.~\ref{fig:violin}, we show the violin plots of these results. Tab.~1 in the supplementary material shows the performance across all datasets using different adaptors and pretrained networks VGG-D,  ResNet18, and ResNet50. Below we summarize the main conclusions.

\noindent
\textbf{Training from scratch is usually not effective.}
Models trained from scratch with synthetic datasets and VGG-D network yield low accuracy. The ResNet18 and ResNet50 networks perform better than VGG-D, but is still quite low compared to using adaptors. This is not surprising as the tasks represent fine-grained classification problems which are quite challenging. On the realistic datasets the performance is better, though when small amounts of data is available the performance is much lower. The results of this approach are shown in the leftmost column of Table~\ref{table:expvgg} and Table 1 of the supplementary material.


\noindent
\textbf{Using initializations for learnable adaptors.} 
For learnable adaptors (linear and multi-layer), we consider random initialization and initialization using a reconstruction. This corresponds to PCA for the linear projection, or as an auto-encoder of the input images with a bottleneck representation equal to three channels as shown in Fig.~\ref{fig:autoencoder}. After training the decoder is discarded and the encoder acts as an adaptor.
As shown in Tab.~\ref{tab:pca}, we find these initializations do not lead to any improvement over a random initialization. 
We suspect that since the adaptor is trained jointly with the rest of the network, the initialization does not impact the performance.

\begin{table}
\centering
\begin{tabular}{l|c c|c c}
\toprule
& \multicolumn{2}{c|}{Linear adaptor} & \multicolumn{2}{c}{Multi-layer adaptor}\\
\cline{2-5}
Dataset & PCA  & Random & Autoenc. & Random \\
\hline
CUB-5   & 38.0 & \textbf{38.4} & \textbf{43.8} & 41.3 \\
Cars-5  & 67.0 & \textbf{70.5} & 72.5 & \textbf{74.4} \\
Aircraft-5 & 78.9 & \textbf{79.6} & 79.3 & \textbf{79.5} \\
So2Sat LCZ42 & \textbf{53.9} & 52.9 & 49.2 & \textbf{52.8} \\
EuroSAT & 94.7 & \textbf{97.1} & 97.5 & \textbf{97.7} \\
LEGUS & 61.0 & \textbf{62.6} & 61.9 & \textbf{62.8}\\
\bottomrule
\end{tabular}
\caption{\textbf{Effect of initialization.} Accuracy using PCA vs. random for linear adaptor, and auto-encoder vs. random initialization for the multi-layer adaptor using VGG-D. \label{tab:pca}}
\end{table}

\noindent
\textbf{Simple adaptors are effective.} When we use VGG-D, simple adaptors like random subset selection and linear adaptors yield the best results. With the exception of EuroSAT, random subset selection and linear projection outperforms all the other single-view adaptors. 
Best results using single-view adaptor networks are shown in green color in Table~\ref{table:expvgg}. 

\begin{figure}[t]
    \centering
    \includegraphics[width=1.0\linewidth]{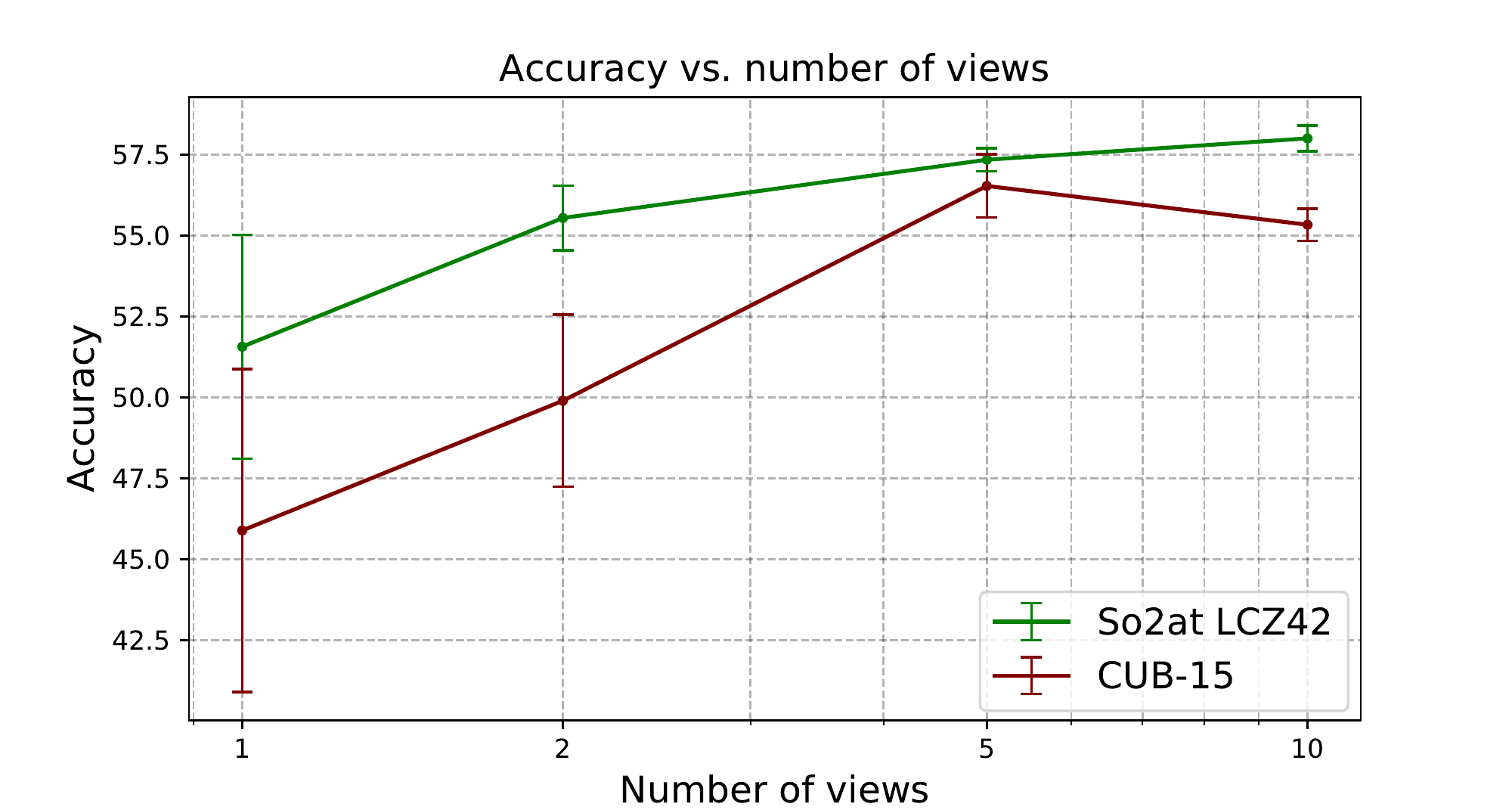}
    \caption{\textbf{Accuracy vs. number of views.} Performance of subset multi-view adaptors varying the number of views. 
    \label{fig:10views}}
\end{figure}

\noindent
\textbf{Multi-view adaptors improve performance.}
Multi-view adaptor networks provide further improvements over single-view for all datasets. For CUB synthetic datasets, we get improvements of up to 11.4\% from the best performing single-view adaptor networks. Using multi-view adaptors also yields an increase of accuracy of up to 4.9\% in the cars dataset and 2.8\% in aircraft. Regarding realistic datasets, we get improvements of up to 2.3\% from the best performing single-view adaptors in So2Sat, 0.2\% in EuroSAT, and 2.1\% in LEGUS (See Fig. \ref{fig:violin}). The best overall results for each dataset are shown in bold blue color in Table~\ref{table:expvgg}.
Fig. \ref{fig:10views} shows accuracies using multi-view adaptors with up to 10 views for So2Sat and CUB-15.
After 5 views the performance does not increase significantly. 

\noindent
\textbf{Shared network for multi-view is effective.}
Combining the information of multiple views increases the performance from the best performing single-view approaches (${\sim}$11\% in some cases). This result occurs despite using a shared architecture to process all the views. The shared network for multi-view adds a similar increase in computational cost as having separate networks for each view compared to single-view approaches. However, the shared network adds a negligible number of parameters to the model.  

\subsection{Domain shift and transferability}
\label{sec:domain_shift}
Here we analyze how the domain shift and the nature of the downstream task impacts transferability.
Since the synthetic images contain the same information as the original color images as the mapping is invertible, the performance difference between the oracle and synthetic dataset indicate the difficulty of transfer using adaptors (see \S~\ref{sec:implementationdetails}). 

On Cars and Aircrafts, adaptors are within 1-2\% of the oracle performance. 
However, on CUB-15 the performance is about 15.3\% lower (72\% $\rightarrow$ 56.7\%).
This difference is likely as cars and aircraft classification can be done using on shape features rather than color, and shape features are largely preserved in the mapping (Fig.~\ref{fig:datasets}). 
This is also indicated by the good performance of the inflated network on these datasets which effectively collapse the color channels.
However, color information is important for bird identification as shape alone is not sufficient. Inflated networks do poorly on this dataset compared to linear adaptors. 

However, one would expect that fine-tuning might be able to ``invert" the mapping that was used to generate the synthetic images. But this appears not to be the case. 
As a further experiment we fine-tuned the network on multiple versions of the CUB dataset where the color channels are permuted (Tab.~\ref{table:permutations}). 
The performance decreases sharply for every permutation except the identity.
The reason for this is that tasks-specific information in color channels is not uniform in natural images (e.g., there are lot more ``red blobs" than ``blue blobs"), which encourages the ImageNet model to learn certain feature detectors better than others (e.g., a ``red blob detector"). 
Thus, mapping the red channel to blue may make a ``red blob" feature ``invisible" to the network. 
While fine-tuning can help, there is a limit due to the phenomenon of critical periods~\cite{achille2017critical} that suggests that initial inhibition of input signal leads to suboptimal performance.

As shown in Tab. \ref{table:degradations}, similar to color permutation, grayscale conversion (which preserves shape rather than color) affects the performance of CUB more than Cars and Aircraft. On the other hand, low resolution (which preserves color rather than shape) reduces the performance of the three datasets similarly. Cross quality distillation \cite{cross-quality-distillation} can be used to alleviate the performance reduction due to image degradation, but it is not a feasible strategy for most realistic datasets.

 \begin{table}
\centering
\begin{tabular}{l|c c c c c c }
\toprule
& \multicolumn{6}{c}{Permutation} \\
\cline{2-7}
Dataset & \textbf{RGB} & RBG & GRB & BGR & GBR & BRG \\
\hline
\multicolumn{7}{c}{VGG-D} \\
\hline
Birds  & \textbf{72.0} & 66.8 & 66.5 & 66.2 & 63.6 & 64.0  \\
Cars & \textbf{78.2} & 78.1 & 78.0 & 77.6 & 77.9 & 76.9  \\
Aircraft & \textbf{81.1} & 81.7 & 81.7 & 82.3 & 81.9 & 82.3  \\
\hline
\multicolumn{7}{c}{ResNet18} \\
\hline
Birds  & \textbf{70.5} & 66.9 & 66.6 & 66.0 & 64.8 & 65.4  \\
Cars & \textbf{80.6} & 80.1 & 80.0 & 79.6 & 79.4 & 79.8  \\
Aircraft & \textbf{76.9} & 76.6 & 76.7 & 76.9 & 76.8 & 76.6  \\
\hline
\multicolumn{7}{c}{ResNet50} \\
\hline
Birds  & \textbf{77.0} & 74.0 & 73.8 & 73.5 & 73.1 & 72.6  \\
Cars & \textbf{86.7} & 86.2 & 85.8 & 85.9 & 85.4 & 85.9  \\
Aircraft & \textbf{84.0} & 83.1 & 83.3 & 83.6 & 82.2 & 83.5  \\
\bottomrule
 \multicolumn{7}{l}{Birds decrease of 7.6\%-11.4\%.} \\ 
 \multicolumn{7}{l}{Cars decrease of 0.9\%-1.2\%.}\\
 \multicolumn{7}{l}{Aircraft decrease of 0.7\%-1.4\%.}\\
\end{tabular}
\caption{\textbf{Accuracy on permuted images.} Accuracy obtained by fine-tuning an ImageNet pretrained VGG-D network on the CUB, Cars, and Aircraft datasets with permuted channels. All permutations except the identity lead to a significant accuracy drop for CUB dataset around $10\%$). In contrast, accuracy in Cars and Aircraft datasets is almost unaffected by the permutations. This is expected because the classification of cars and aircraft depends mostly on shape than color.  
\label{table:permutations}}
\end{table}

 \begin{table}
 \renewcommand{\arraystretch}{1.01}
\centering
\resizebox{\columnwidth}{!}{
\begin{tabular}{l c|c c|c c|c c}
\toprule
        &               & \multicolumn{2}{c|}{Channel}     &   &        & \multicolumn{2}{c}{Low}   \\
Dataset & \textbf{RGB}  & \multicolumn{2}{c|}{permutation} & \multicolumn{2}{c|}{Grayscale} & \multicolumn{2}{c}{resolution}  \\
\hline
\multicolumn{8}{c}{VGG-D} \\
\hline
Birds  & \textbf{72.0} & 65.4 & (11.4\%) & 54.5 & (24.3\%) & 58.4 & (18.9\%)  \\
Cars & \textbf{78.2} & 77.7 & (1.0\%) & 76.5 & (2.2\%) & 61.0 & (22.0\%)  \\
Aircraft & \textbf{81.1} & 82.0 & (-1.3\%) & 80.0 & (1.3\%) & 72.4 & (10.7\%)  \\
\hline
\multicolumn{8}{c}{ResNet18} \\
\hline
Birds  & \textbf{70.5} & 65.9 & (7.7\%) & 50.9 & (27.8\%) & 56.1 & (20.4\%)  \\
Cars & \textbf{80.6} & 79.8 & (1.2\%) & 78.8 & (2.3\%)  & 68.0 & (15.7\%)   \\
Aircraft & \textbf{76.9} & 76.7 & (0.3\%)  & 75.2 & (2.3\%)  & 68.3 & (11.2\%)   \\
\hline
\multicolumn{8}{c}{ResNet50} \\
\hline
Birds  & \textbf{77.0} & 73.4 & (5.4\%)  & 61.5 & (20.1\%)  & 66.7 & (13.4\%)   \\
Cars & \textbf{86.7} & 85.8 & (1.2\%)  & 85.5 & (1.4\%)  & 76.1 & (12.2\%)   \\
Aircraft & \textbf{84.0} & 84.0 & (1.4\%)  & 82.2 & (2.1\%)  & 77.7 & (7.5\%)   \\
\bottomrule
\end{tabular}}
\caption{\textbf{Accuracy with image degradations}. Similar to color permutation, grayscale conversion (which preserves shape rather than color) affects the performance of CUB dataset in a more significant way than Cars and Aircraft. On the other hand, low resolution (which preserves color rather than shape) reduces the performance of the three datasets similarly. We show the percentage decrease in parenthesis.
\label{table:degradations}}
\end{table}


These experiments suggest that there is a limit to transferability using adaptors and the final performance (in the limit of large training dataset) may be worse than training from scratch when large labeled datasets are available. However, transfer learning with adaptors is practical for the few-shot setting. 
It also suggests that alternate optimization techniques (e.g., discrete search over permutations) might be more effective for learning adaptors than gradient based optimization.

While we do not know a similar ``upper bound" in accuracy for the realistic datasets, we include two experiments on EuroSAT and LEGUS to provide a sense of the task difficulty.

\noindent
\textbf{Hyperspectral vs. color EuroSAT.} EuroSAT provides an RGB version of the images in addition to the 13 spectral bands. We use these to fine-tune the pretrained model and compare the performance using adaptors trained on the 13 channels. 
We obtain 70.3\% accuracy with VGG-D network when training on RGB images compared to 97.7\% accuracy on the hyperspectral images. 
This is no surprise since these hyperspectral channels were chosen for their informativeness on the tasks.

\noindent
\textbf{Comparison to a manual adaptor for LEGUS.}
Similar to the EuroSAT dataset, we fine-tune the pretrained network using LEGUS images converted to RGB using a conventional method used to visualize astronomical images. The three channels are constructed by merging the first two and the last two channels as:
\begin{itemize}
\setlength\itemsep{0.1em}
    \item R $= (\gamma_v V + \gamma_i I)/2$\vspace{-3pt}
    \item G $= (\gamma_b B)$\vspace{-3pt}
    \item B $= (\gamma_n NUV + \gamma_uU)/2$
\end{itemize}
where $\gamma$ is the inverse gain of the filters used by the HST to capture the images. We obtain 62.8\% accuracy with a VGG-D network using these RGB images. In contrast, when using the spectral bands separately we achieve 65.3\% accuracy.

\section{Conclusion}
\label{sec:conclusion}
We considered the problem of adapting a color image network to a hyperspectral domain by designing lightweight adaptors. 
We found that simple schemes such as linear projection and subset selection are effective, but the performance can be low when the number of channels are large. 
Multi-view adaptors provide consistent gains in such cases and are a compelling alternative to ensembles due to the shared model structure.
We also highlighted some difficulties in transfer learning with gradient-based fine-tuning of adaptors caused due to domain mismatch.
Alternate parameterizations of adaptors (e.g., using transformers~\cite{vaswani2017attention}) or search based optimization could alleviate this problem and is a subject to our future work.
Future work will also investigate techniques from the self-supervised and semi-supervised learning to improve transfer when domain shifts are large.



\ifCLASSOPTIONcaptionsoff
  \newpage
\fi




\begin{thebibliography}{1}

\bibitem{achille2017critical}
A. Achille, M. Rovere, and S. Soatto. \emph{Critical learning periods in deep neural networks}. arXiv preprint arXiv:1711.08856, 2017.

\bibitem{barchi2020}
P.H. Barchi, R.R. de Carvalho, R.R. Rosa, R.A. Sautter, M. Soares-Santos, B.A.D. Marques, E. Clua, T.S. Gonc¸alves, C. de S´a-Freitas, and T.C. Moura. \emph{Machine and deep learning applied to galaxy morphology - a comparative study}. Astronomy and Computing, 30:100334, Jan 2020.

\bibitem{Bialopetravicius2020}
J. Bialopetravičius and D. Narbutis. \emph{Deriving star cluster parameters with convolutional neural networks}. Astronomy Astrophysics, 633:A148, Jan 2020.

\bibitem{Bialopetravicius2019}
J. Bialopetravičius, D. Narbutis, and V. Vansevičius. \emph{Deriving star cluster parameters with convolutional neural networks}. Astronomy Astrophysics, 621:A103, Jan 2019.

\bibitem{legus}
D. Calzetti, J. Lee, E. Sabbi, and A. Adamo. \emph{Legacy extragalactic uv survey (legus) with the hubble space telescope. i. survey description}. The Astronomical Journal, 149(51):25, 2015.

\bibitem{camps2018}
G. Camps-Valls, D. Tuia, L. Bruzzone, J. A. Benediktsson. \emph{Advances in Hyperspectral Image Classification: Earth monitoring with statistical learning methods}. CoRR, 2013.
               
\bibitem{medimg}
O. Carrasco, R. B. Gomez, A. Chainani, and W. E. Roper. \emph{Hyperspectral imaging applied to medical diagnoses and food safety}. In Geo-Spatial and Temporal Image and Data Exploitation III, volume 5097, pages 215 – 221. International Society for Optics and Photonics, SPIE, 2003.

\bibitem{inflated}
J. Carreira and A. Zisserman. \emph{Quo vadis, action recognition? A new model and the kinetics dataset}. CoRR, abs/1705.07750, 2017.

\bibitem{imagenet}
J. Deng,  W. Dong,  R. Socher,  L.-J. Li,  K. Li,  and L. Fei-Fei. \emph{ImageNet: A Large-Scale Hierarchical Image Database}. In Proceedings of the IEEE Conference on Computer Vision and Pattern Recognition (CVPR), 2009.

\bibitem{Ferris2001}
D. Ferris, R. Lawhead, E. D. Dickman, N. Holtzapple, J. A. Miller, S. Grogan, S. Bambot, A. Agrawal, and M. Faupel. \emph{Multimodal hyperspectral imaging for the noninvasive diagnosis of cervical neoplasia}. Journal of Lower Genital Tract Disease, 5:65–72, 2001.

\bibitem{goodfellow2014generative}
I. Goodfellow, J. Pouget-Abadie, M. Mirza, B. Xu, D. Warde-Farley, S. Ozair, A. Courville, and
Y. Bengio. \emph{Generative adversarial nets}. In Advances in neural information processing systems, pages 2672–2680, 2014.

\bibitem{goyal}
P. Goyal, P. Dollar, R. B. Girshick, P. Noordhuis, L. Wesolowski, A. Kyrola, A. Tulloch, Y. Jia, and K. He. \emph{Accurate, large minibatch SGD: training imagenet in 1 hour}. CoRR, abs/1706.02677,
2017.

\bibitem{gupta2014}
S. Gupta, R. Girshick, P. Arbelaez, and J. Malik. \emph{Learning rich features from rgb-d images for object detection and segmentation}. CoRR, 2014.

\bibitem{morpheus}
R. Hausen and B. E. Robertson. \emph{Morpheus: A deep learning framework for the pixel-level analysis of astronomical image data}. The Astrophysical Journal Supplement Series, 248(1):20, May 2020.

\bibitem{resnet}
K. He, X. Zhang, S. Ren, and J. Sun. \emph{Deep residual learning for image recognition}. CoRR, abs/1512.03385, 2015.

\bibitem{LC3}
L. He, J. Li, A. Plaza, and Y. Li. \emph{Discriminative low-rank gabor filtering for spectral-spatial hyperspectral image classification}. IEEE Transactions on Geoscience and Remote Sensing, PP:1–15, 12 2016.

\bibitem{eurosat}
P. Helber, B. Bischke, A. Dengel, and D. Borth. \emph{Eurosat: A novel dataset and deep learning benchmark for land use and land cover classification}. CoRR, abs/1709.00029, 2017.

\bibitem{herath2019}
S. Herath, M. Harandi, B. Fernando, and R. Nock. \emph{Min-max statistical alignment for transfer learning}. In Proceedings of the IEEE/CVF Conference on Computer Vision and Pattern Recognition (CVPR), June 2019.

\bibitem{houlsby2019}
N. Houlsby, A. Giurgiu, S. Jastrzebski, B. Morrone, Q. de Laroussilhe, A. Gesmundo, M. Attariyan, and S. Gelly. \emph{Parameter-efficient transfer learning for nlp}. CoRR, 2019.

\bibitem{huang2020}
Y. Huang, Y. Guo, and C. Gao. \emph{Efficient parallel inflated 3d convolution architecture for action recognition}. IEEE Access, 8:45753–45765, 2020.

\bibitem{bn}
S. Ioffe and C. Szegedy. \emph{Batch normalization: Accelerating deep network training by reducing internal covariate shift}. CoRR, abs/1502.03167, 2015.

\bibitem{pca}
I.T. Jolliffe. \emph{Principal Component Analysis}. Springer Verlag, 1986.

\bibitem{khalifa2017}
N. M. Khalifa, M. N. Taha, A. Hassanien, and I. M. Selim. \emph{Deep galaxy: Classification
of galaxies based on deep convolutional neural networks}. CoRR, 2017.

\bibitem{LC1}
M. Khodadadzadeh, J. Li, A. Plaza, H. Ghassemian, J. Bioucas-Dias, and X. Li. \emph{Spectral–spatial classification of hyperspectral data using local and global probabilities for mixed pixel characterization}. Geoscience and Remote Sensing, IEEE Transactions on, 52:6298–6314, 10 2014.

\bibitem{adam}
D. P. Kingma and J. Ba. \emph{Adam: A method for stochastic optimization}. Published as a conference paper at the 3rd International Conference for Learning Representations, San Diego, 2015.

\bibitem{kornblith2019}
S. Kornblith, J. Shlens, and Q. V. Le. \emph{Do better imagenet models transfer better?}. In Proceedings of the IEEE/CVF Conference on Computer Vision and Pattern Recognition (CVPR), June 2019.

\bibitem{cars}
J. Krause, M. Stark, J. Deng, and L. Fei-Fei. \emph{3d object representations for fine-grained categorization}. In 4th International IEEE Workshop on 3D Representation and Recognition (3dRR-13), Sydney, Australia, 2013.

\bibitem{lalonde2019}
R. LaLonde, I. Tanner, K. Nikiforaki, G. Z. Papadakis, P. Kandel, C. W. Bolan, M. B. Wallace, and U. Bagci. \emph{Inn: Inflated neural networks for ipmn diagnosis}. Medical Image Computing and Computer Assisted Intervention (MICCAI), 2019.

\bibitem{bcnn}
T. Lin, A. RoyChowdhury, and S. Maji. \emph{Bilinear cnns for fine-grained visual recognition}. In Transactions of Pattern Analysis and Machine Intelligence (PAMI), 2017.

\bibitem{long2017}
M. Long, H. Zhu, J. Wang, and M. I. Jordan. \emph{Deep transfer learning with joint adaptation networks}. CoRR, 2017.

\bibitem{aircrafts}
S. Maji, J. Kannala, E. Rahtu, M. Blaschko, and A. Vedaldi. \emph{Fine-grained visual classification of aircraft}. Technical report, CoRR, 2013.

\bibitem{martin2019}
G Martin, S Kaviraj, A Hocking, S C Read, and J E Geach. \emph{Galaxy morphological classification in deep-wide surveys via unsupervised machine learning}. Monthly Notices of the Royal Astronomical Society, 491(1):1408–1426, Oct 2019.

\bibitem{mishra2006}
A. K. Mishra and B. Mulgrew. \emph{Radar signal classification using pca-based features}. In 2006 IEEE International Conference on Acoustics Speech and Signal Processing Proceedings, volume 3, pages III–III, 2006.

\bibitem{mittal2019}
A. Mittal, A. Soorya, P. Nagrath, and D. J. Hemanth. \emph{Data augmentation based morphological classification of galaxies using deep convolutional neural network}. Earth Science Informatics, 13:601 – 617, 2019.

\bibitem{qiu2020}
C. Qiu, X. Tong, M. Schmitt, B. Bechtel, and X. X. Zhu. \emph{Multilevel feature fusion-based cnn for local climate zone classification from sentinel-2 images: Benchmark results on the so2sat lcz42 dataset}. IEEE Journal of Selected Topics in Applied Earth Observations and Remote Sensing, 13:2793– 2806, 2020.

\bibitem{radosavovic2017}
I. Radosavovic, P. Dollar, R. Girshick, G. Gkioxari, and K. He. \emph{Data distillation: Towards omnisupervised learning}. CoRR, 2017.

\bibitem{fasterrcnn}
S. Ren, K. He, R. Girshick, and J. Sun. \emph{Faster r-cnn: Towards real-time object detection with region proposal networks}. In Advances in Neural Information Processing Systems, volume 28. Curran Associates, Inc., 2015.

\bibitem{vgg}
K. Simonyan and A. Zisserman. \emph{Very deep convolutional networks for large-scale image recognition}. International Conference on Learning Representations (ICLR), 2015.

\bibitem{simonyan2014}
K. Simonyan and A. Zisserman. \emph{Two-Stream Convolutional Networks for Action Recognition in Videos}.  Advances in Neural Information Processing Systems (NIPS), 2014.

\bibitem{multi-view}
H. Su, S. Maji, E. Kalogerakis, and E. Learned-Miller. \emph{Multi-view convolutional neural networks for 3d shape recognition}. In Proceedings of the 2015 IEEE International Conference on Computer Vision, ICCV ’15, page 945–953, USA, 2015. IEEE Computer Society.

\bibitem{cross-quality-distillation}
J. Su and S. Maji. Cross quality distillation. CoRR, abs/1604.00433, 2016.

\bibitem{vaswani2017attention}
A. Vaswani, N. Shazeer, N. Parmar, J. Uszkoreit, L. Jones, A. N. Gomez, L. Kaiser, and I. Polosukhin. \emph{Attention is all you need}. arXiv preprint arXiv:1706.03762, 2017.

\bibitem{cub}
C. Wah, S. Branson, P. Welinder, P. Perona, and S. Belongie. \emph{The Caltech-UCSD Birds-200-2011 Dataset}. Technical report, 2011.

\bibitem{wang2020}
J. Wang, Y. Chen, W. Feng, H. Yu, M. Huang, and Q. Yang. \emph{Transfer learning with dynamic distribution adaptation}. ACM Transactions on Intelligent Systems and Technology, 11(1):1–25, Feb 2020.

\bibitem{LC4}
Q. Wang, Z. Meng, and X. Li. \emph{Locality adaptive discriminant analysis for spectral-spatial classification of hyperspectral images}. IEEE Geoscience and Remote Sensing
Letters, PP:1–5, 09 2017.

\bibitem{wei2020}
W. Wei, E. A. Huerta, B. C. Whitmore, J. C. Lee, S. Hannon, R. Chandar, D. A. Dale, K. L. Larson, D. A. Thilker, L. Ubeda, M. Boquien, M. Chevance, J. M. D. Kruijssen, A. Schruba, G. A. Blanc, and E. Congiu. \emph{Deep transfer learning for star cluster classification: I. application to the PHANGS-HST survey}. , 493(3):3178–3193, Feb. 2020.

\bibitem{weiss2016}
K. Weiss, T. M. Khoshgoftaar, and D. Wang. \emph{A survey of transfer learning}. Journal of Big data, 3(1):9, 2016.

\bibitem{LC2}
X. Xu, J. Li, X. Huang, M. D. Mura, and A. Plaza. \emph{Multiple morphological component analysis based decomposition for remote sensing image classification}. IEEE Transactions on Geoscience and Remote Sensing, 54:1–20, 01 2016.

\bibitem{transferlearning}
J. Yosinski, J. Clune, Y. Bengio, and H. Lipson. \emph{How transferable are features in deep neural networks?}. In Advances in Neural Information Processing Systems 27, pages 3320–3328. Curran Associates, Inc., 2014.

\bibitem{so2sa}
X. Zhu, J. Hu, C. Qiu, Y. Shi, H. Bagheri, J. Kang, H. Li, L. Mou, G. Zhang, M. Haberle, S. Han, Y. Hua, R. Huang, L. Hughes, Y. Sun, M. Schmitt, and Y. Wang. \emph{So2sat lcz42}, 2018.

\bibitem{zhu2019}
X. Zhu, J. Dai, C. Bian, Y. Chen, S. Chen, and C. Hu. \emph{Galaxy morphology classification with deep convolutional neural networks}. Astrophysics and Space Science, 364(4), Apr 2019.

\bibitem{zhu2017}
X. X. Zhu, D. Tuia, L. Mou, G. Xia, L. Zhang, F. Xu, and F. Fraundorfer. \emph{Deep learning in remote sensing: A comprehensive review and list of resources}. IEEE Geoscience and Remote Sensing Magazine, 5(4):8–36, 2017.

\bibitem{zhuang2020}
F. Zhuang, Z. Qi, K. Duan, D. Xi, Y. Zhu, H. Zhu, H. Xiong, and Q. He. \emph{A comprehensive survey on transfer learning}. CoRR, 2020.


\end{thebibliography}
%

%
\begin{IEEEbiographynophoto}{Gustavo P\'erez}
is a PhD student in the College of Information and Computer Sciences at the University of Massachusetts, Amherst. He received a M.Sc. in Biomedical Engineering at Universidad de los Andes, and a B.Eng. in Electronic Engineering from Universidad del Norte in Colombia. 
\end{IEEEbiographynophoto}
\begin{IEEEbiographynophoto}{Subhransu Maji}
is an Associate Professor in the College of Information and Computer Sciences at the University of Massachusetts, Amherst. He obtained his PhD from the University of California at Berkeley, and a B.Tech. in Computer Science and Engineering from IIT Kanpur. 
\end{IEEEbiographynophoto}






\clearpage\newpage
\begin{table*}[t]
\renewcommand{\arraystretch}{1.01}
\centering
\resizebox{\textwidth}{!}{%
\begin{tabular}{l c|c c c c|c c c c}
  \multicolumn{10}{c}{\textbf{\Large Supplementary Material}}\\
  \multicolumn{10}{c}{}\\
  \multicolumn{10}{c}{\textbf{(a) Performance using VGG-D network}}\\
\toprule
& & \multicolumn{4}{c|}{Single-view} &  \multicolumn{4}{c}{Multi-view} \\
        \cline{3-10}
          & From & Linear & Inflated & Multi-layer &  Subset   & \multicolumn{2}{c}{Subset selection} & \multicolumn{2}{c}{Linear adaptor}      \\
Dataset    & scratch & adaptor & network & adaptor & selection & 2 & 5 & 2 & 5 \\
\hline
\emph{Synthetic}\\
\hline
CUB-5       & $0.5\pm0.0$ & $38.4\pm3.1$ & $30.5\pm1.1$ & $41.3\pm2.8$ & $\best{47.1\pm4.0}$  
            & $45.0\pm1.7$ & $49.8\pm0.5$ & $52.7\pm0.6$ & $\bestm{55.1\pm0.3}$  \\
CUB-15      & $0.5\pm0.0$ & $41.2\pm0.9$ & $26.0\pm2.8$ & $\best{45.4\pm5.1}$ & $\best{45.9\pm5.0}$  
            & $49.9\pm2.7$ & $\bestm{56.5\pm1.0}$ & $\bestm{57.3\pm0.5}$ & $\bestm{56.7\pm0.9}$  \\
\hline
Cars-5      & $2.2\pm0.8$ & $70.5\pm2.1$ & $70.2\pm1.4$ & $\best{74.4\pm1.7}$ & $\best{72.6\pm2.0}$  
            & $75.4\pm0.9$ & $\bestm{76.7\pm0.5}$ & $\bestm{76.7\pm0.4}$ & $\bestm{76.8\pm0.5}$   \\
Cars-15     & $2.7\pm0.4$ & $\best{70.7\pm3.3}$ & $68.7\pm3.4$ & $\best{70.1\pm2.3}$ & $\best{72.2\pm1.7}$  
            & $75.8\pm0.4$ & $77.1\pm0.3$ & $77.1\pm0.2$ & $\bestm{78.1\pm0.4}$   \\
 \hline
\hline
Aircraft-5  &  $1.0\pm0.0$ & $\best{79.6\pm0.8}$ & $76.6\pm0.9$ & $\best{79.5\pm0.5}$ & $\best{79.2\pm1.0}$  
            & $\bestm{80.8\pm0.9}$ & $\bestm{81.5\pm0.3}$ & $81.1\pm0.4$ & $\bestm{81.8\pm0.2}$   \\
Aircraft-15 &  $1.0\pm0.0$ & $78.4\pm0.6$ & $77.6\pm0.9$ & $79.1\pm1.2$ & $\bestm{79.8\pm2.2}$  
            & $\bestm{81.3\pm1.0}$ & $\bestm{81.6\pm0.5}$ & $\bestm{81.4\pm0.5}$ & $\bestm{81.3\pm0.4}$    \\
\hline
\emph{Realistic}\\
\hline
So2Sat     & $50.3\pm0.8$ & $\best{52.8\pm0.3}$ & $51.8\pm0.3$ & $\best{52.8\pm0.7}$ & $\best{51.6\pm3.5}$ 
          & $55.6\pm1.0$ & $\bestm{57.4\pm0.4}$ & $\bestm{57.0\pm0.4}$ & $\bestm{57.2\pm0.5}$    \\
So2Sat$^S$ & $34.5\pm0.7$ & $\best{45.1\pm0.5}$ & $40.8\pm0.8$ & $40.8\pm3.1$ & $\best{43.8\pm4.9}$  
          & $49.8\pm1.2$ & $\bestm{51.4\pm0.3}$ & $45.2\pm0.7$ & $48.6\pm0.8$   \\
\hline
EuroSAT     & $95.7\pm0.2$ & $97.1\pm0.3$ & $96.7\pm0.3$ & $\best{97.7\pm0.2}$ & $96.6\pm0.7$  
            & $\bestm{97.4\pm0.4}$ & $\bestm{97.7\pm0.2}$ & $\bestm{97.4\pm0.2}$ & $\bestm{97.5\pm0.2}$   \\
EuroSAT$^S$ & $75.2\pm2.6$ & $88.4\pm0.5$ & $83.6\pm1.8$ & $\best{92.8\pm0.7}$ & $86.6\pm3.6$  
            & $91.6\pm2.2$ & $93.7\pm0.5$ & $\bestm{94.4\pm0.2}$ & $\bestm{94.4\pm0.2}$  \\
\hline
LEGUS       & $51.9\pm0.5$ & $\best{63.2\pm0.5}$ & $61.4\pm0.4$ & $\best{62.8\pm0.3}$ & $61.1\pm1.4$  
            & $64.2\pm0.4$ & $\bestm{65.3\pm0.5}$ & $64.2\pm0.4$ & $63.9\pm0.6$   \\
LEGUS$^S$   & $27.2\pm5.5$ & $\best{54.9\pm0.8}$ & $51.8\pm1.3$ & $\best{53.1\pm1.9}$ & $\best{51.3\pm2.9}$  
            & $55.0\pm0.6$ & $\bestm{59.3\pm0.6}$ & $55.7\pm1.8$ & $\bestm{60.0\pm1.6}$    \\
 \bottomrule
  \multicolumn{10}{c}{\textbf{(b) Performance using ResNet18 network}}\\
\toprule
     & & \multicolumn{4}{c|}{Single-view} &  \multicolumn{4}{c}{Multi-view} \\
        \cline{3-10}
          & From & Linear & Inflated & Multi-layer & Subset   & \multicolumn{2}{c}{Subset selection} & \multicolumn{2}{c}{Linear adaptor}      \\
Dataset    & scratch & adaptor & network & adaptor & selection & 2 & 5 & 2 & 5 \\
\hline
\emph{Synthetic}\\
\hline
CUB-5        & $12.0\pm0.4$ & $\best{46.0\pm0.9}$ & $41.1\pm0.7$ & $\best{46.1\pm2.3}$ & $\best{48.1\pm2.8}$ 
             & $48.5\pm1.3$ & $47.6\pm1.0$ & $\bestm{58.1\pm0.5}$ & $55.8\pm0.3$  \\
CUB-15       & $15.7\pm0.7$ & $\best{50.9\pm1.6}$ & $40.8\pm2.4$ & $\best{47.6\pm2.7}$ & $\best{51.6\pm4.6}$ 
             & $50.3\pm4.2$ & $51.7\pm3.1$ & $\bestm{60.5\pm1.8}$ & $\bestm{57.3\pm2.6}$  \\
\hline
Cars-5        & $9.1\pm0.5$ & $\bestm{75.4\pm0.4}$ & $72.6\pm0.7$ & $\bestm{76.7\pm1.4}$ & $73.6\pm1.4$ 
              & $73.9\pm1.4$ & $74.2\pm0.2$ & $\bestm{75.1\pm0.5}$ & $\bestm{75.3\pm0.4}$  \\
Cars-15       & $15.7\pm0.9$ & $73.8\pm0.4$ & $72.3\pm0.4$ & $\bestm{77.0\pm1.3}$ & $\bestm{74.5\pm1.4}$ 
              & $74.6\pm1.2$ & $74.6\pm0.4$ & $\bestm{76.4\pm0.3}$ & $\bestm{76.4\pm0.6}$  \\
\hline
Aircraft-5  &  $38.1\pm0.8$ & $73.3\pm0.4$ & $73.4\pm0.4$ & $\bestm{75.6\pm0.9}$ & $\bestm{74.3\pm1.1}$  
            & $\bestm{74.0\pm1.1}$ & $73.8\pm0.3$ & $73.6\pm0.6$ & $\bestm{75.1\pm0.6}$   \\
Aircraft-15 &  $44.1\pm0.6$ & $73.3\pm0.4$ & $73.9\pm0.4$ & $\bestm{75.8\pm0.7}$ & $\bestm{74.9\pm0.8}$  
            & $73.9\pm0.4$ & $74.0\pm0.7$ & $\bestm{74.1\pm1.3}$ & $73.3\pm1.4$    \\
\hline
\emph{Realistic}\\
\hline
So2Sat         & $\best{47.9\pm0.3}$ & $\best{47.6\pm0.8}$ & $46.7\pm0.9$ & $\best{48.7\pm0.7}$ & $\best{48.1\pm2.7}$ 
              & $52.8\pm1.0$ & $\bestm{53.9\pm1.0}$ & $52.4\pm1.1$ & $\bestm{54.2\pm0.6}$  \\
So2Sat$^S$     & $36.1\pm0.5$ & $\best{41.7\pm1.0}$ & $38.0\pm0.8$ & $\best{40.4\pm1.6}$ & $\best{41.5\pm3.9}$ 
              & $45.3\pm1.1$ & $\bestm{48.3\pm0.1}$ & $43.3\pm1.0$ & $46.6\pm0.5$  \\
\hline
EuroSAT         & $97.7\pm0.0$ & $\best{98.4\pm0.1}$ & $\best{98.4\pm0.1}$ & $\best{98.2\pm0.1}$ & $97.7\pm0.5$ 
                & $\bestm{98.3\pm0.1}$ & $\bestm{98.5\pm0.2}$ & $98.1\pm0.1$ & $\bestm{98.3\pm0.1}$  \\
EuroSAT$^S$     & $90.4\pm0.3$ & $94.0\pm0.3$ & $\best{95.4\pm0.2}$ & $93.9\pm0.4$ & $\best{94.0\pm0.9}$ 
                & $95.3\pm0.3$ & $\bestm{96.2\pm0.2}$ & $95.1\pm0.1$ & $\bestm{96.0\pm0.1}$  \\
\hline
LEGUS       & $57.8\pm2.0$ & $\best{64.2\pm0.6}$ & $62.5\pm0.7$ & $\best{63.4\pm0.6}$ & $61.0\pm3.3$ 
            & $\bestm{65.7\pm0.6}$ & $\bestm{65.1\pm0.3}$ & $64.4\pm0.0$ & $\bestm{65.1\pm0.6}$  \\
LEGUS$^S$   & $48.4\pm0.4$ & $52.6\pm0.3$ & $\best{55.3\pm0.8}$ & $52.0\pm1.0$ & $48.7\pm1.6$ 
            & $\bestm{57.8\pm0.4}$ & $\bestm{58.1\pm0.8}$ & $56.4\pm0.5$ & $\bestm{57.9\pm0.3}$  \\
 \bottomrule
  \multicolumn{10}{c}{\textbf{(c) Performance using ResNet50 network}}\\
\toprule
     & & \multicolumn{4}{c|}{Single-view} &  \multicolumn{4}{c}{Multi-view} \\
        \cline{3-10}
          & From & Linear & Inflated & Multi-layer & Subset   & \multicolumn{2}{c}{Subset selection} & \multicolumn{2}{c}{Linear adaptor}      \\
Dataset    & scratch & adaptor & network & adaptor & selection & 2 & 5 & 2 & 5 \\
\hline
\emph{Synthetic}\\
\hline
CUB-5        & $12.3\pm0.2$ & $\best{61.7\pm0.6}$ & $55.7\pm0.8$ & $\best{61.3\pm3.0}$ & $\best{59.7\pm1.8}$  
             & $59.6\pm1.0$ & $58.3\pm1.2$ & $\bestm{66.4\pm0.2}$ & $64.8\pm0.2$  \\
CUB-15       & $13.8\pm0.2$ & $\best{67.9\pm0.8}$ & $57.4\pm0.5$ & $63.1\pm2.6$ & $62.2\pm3.8$  
             & $62.9\pm3.3$ & $63.9\pm1.3$ & $\bestm{68.2\pm0.2}$ & $66.8\pm1.7$  \\
\hline
Cars-5        & $12.8\pm0.6$ & $\bestm{83.2\pm0.5}$ & $82.4\pm0.3$ & $\bestm{83.8\pm0.3}$ & $82.8\pm0.6$ 
              & $82.5\pm0.9$ & $82.1\pm0.5$ & $82.2\pm0.3$ & $\bestm{83.3\pm0.4}$  \\
Cars-15       & $14.6\pm1.2$ & $83.4\pm0.2$ & $82.0\pm0.4$ & $\bestm{84.5\pm0.8}$ & $82.8\pm0.7$ 
              & $\bestm{83.2\pm0.6}$ & $82.3\pm0.3$ & $\bestm{83.9\pm0.4}$ & $\bestm{84.0\pm0.5}$  \\
\hline
Aircraft-5  &  $34.2\pm0.9$ & $81.6\pm0.3$ & $81.3\pm0.5$ & $\bestm{83.0\pm1.0}$ & $81.8\pm0.1$  
            &  $81.2\pm0.6$ & $80.5\pm0.5$ & $80.9\pm0.7$ & $80.4\pm0.6$   \\
Aircraft-15 &  $38.6\pm1.0$ & $\bestm{81.8\pm0.4}$ & $\bestm{81.7\pm0.3}$ & $\bestm{82.4\pm0.5}$ & $\bestm{82.0\pm0.7}$  
            &  $81.0\pm0.3$ & $80.7\pm0.5$ & $81.0\pm0.5$ & $\bestm{81.4\pm0.6}$    \\
\hline
\emph{Realistic}\\
\hline
So2Sat         & $46.5\pm0.5$ & $47.7\pm1.1$ & $\best{50.6\pm1.0}$ & $\best{50.7\pm1.3}$ & $\best{47.7\pm3.0}$ 
              & $52.6\pm0.6$ & $\bestm{54.9\pm0.6}$ & $52.2\pm1.1$ & $\bestm{54.6\pm0.7}$  \\
So2Sat$^S$     & $32.0\pm1.6$ & $38.1\pm0.5$ & $\best{42.9\pm1.1}$ & $\best{44.8\pm0.9}$ & $41.3\pm1.7$ 
              & $\bestm{48.9\pm1.4}$ & $\bestm{50.5\pm0.2}$ & $44.4\pm0.7$ & $49.1\pm0.9$  \\
\hline
EuroSAT         & $97.1\pm0.1$ & $\bestm{98.6\pm0.1}$ & $\bestm{98.6\pm0.1}$ & $\best{98.3\pm0.2}$ & $98.0\pm0.3$ 
                & $\bestm{98.5\pm0.2}$ & $\bestm{98.7\pm0.1}$ & $98.1\pm0.1$ & $98.4\pm0.1$  \\
EuroSAT$^S$     & $80.3\pm0.8$ & $\best{95.6\pm0.2}$ & $\best{95.5\pm0.1}$ & $\best{94.8\pm0.7}$ & $\best{94.6\pm1.0}$ 
                & $\bestm{96.2\pm0.4}$ & $\bestm{96.5\pm0.2}$ & $95.7\pm0.3$ & $\bestm{96.4\pm0.1}$  \\
\hline
LEGUS       & $62.6\pm0.4$ & $\best{64.1\pm0.4}$ & $\best{63.9\pm0.4}$ & $\bestm{64.3\pm0.5}$ & $\bestm{63.2\pm1.8}$ 
            & $\bestm{65.1\pm0.4}$ & $\bestm{65.5\pm0.7}$ & $\bestm{64.9\pm0.5}$ & $\bestm{65.0\pm0.6}$  \\
LEGUS$^S$   & $49.4\pm1.8$ & $\best{55.5\pm0.7}$ & $\best{56.5\pm0.7}$ & $54.1\pm0.6$ & $53.7\pm1.5$ 
            & $57.4\pm1.4$ & $\bestm{59.2\pm0.8}$ & $58.1\pm0.6$ & $\bestm{60.3\pm0.8}$  \\
 \bottomrule
 \multicolumn{5}{l}{\footnotesize $^S$ Smaller version of the dataset using 1000 training samples.}\\
\end{tabular}}
\caption{\textbf{Results using hyperspectral domain adaptation methods.} Accuracy (\%) for single and multi-view adaptor networks on the six proposed datasets using VGG-D, ResNet18, and ResNet50. 
We present results of the six synthetic datasets (CUB-5, CUB-15, Cars-5, Cars-15, Aircraft-5, and Aircraft-15) in the first six rows. Also, we include results of the three realistic datasets (and its reduced variants) in the last six rows for each architecture. We show results of the baselines (From Scratch), of the four adaptors, and results using our multi-view scheme for random subset selection and linear adaptor. The best results using single-view adaptor networks are shown in green color. The best overall results are shown in bold blue color.
\label{table:expall}}
\end{table*}

\end{document}